\documentclass[letterpaper]{article} 
\usepackage{aaai2026}  
\usepackage{times}  
\usepackage{helvet}  
\usepackage{courier}  
\usepackage[hyphens]{url}  
\usepackage{graphicx} 
\usepackage{natbib}  
\usepackage{caption} 
\usepackage{algorithm}
\usepackage{algorithmic}

\usepackage{booktabs}
\usepackage{multicol}
\usepackage{multirow}
\usepackage{tabularx}

\usepackage{newfloat}
\usepackage{listings}
\DeclareCaptionStyle{ruled}{labelfont=normalfont,labelsep=colon,strut=off} 
\floatstyle{ruled}
\newfloat{listing}{tb}{lst}{}
\floatname{listing}{Listing}
\title{Assessing Electricity Demand Forecasting with Exogenous Data in Time Series Foundation Models}
\author {
    Wei Soon Cheong\equalcontrib,
    Lian Lian Jiang\equalcontrib\thanks{Corresponding Author.},
    Jamie Ng Suat Ling
}
\affiliations {
    Institute for Infocomm Research (I\textsuperscript{2}R), Agency for Science, Technology and Research (A*STAR)\\
    1 Fusionopolis Way, 21-01 Connexis, 138632, Singapore\\
    caedmoncheong@gmail.com, 
    Jiang\_Lianlian@a-star.edu.sg\textsuperscript{\textdagger},
    Jamie@a-star.edu.sg
}

\begin{document}

\maketitle

\begin{abstract}
Time-series foundation models have emerged as a new paradigm for forecasting, yet their ability to effectively leverage exogenous features---critical for electricity demand forecasting---remains unclear. This paper empirically evaluates foundation models capable of modeling cross-channel correlations against a baseline LSTM with reversible instance normalization across Singaporean and Australian electricity markets at hourly and daily granularities. We systematically assess MOIRAI, MOMENT, TinyTimeMixers, ChronosX, and Chronos-2 under three feature configurations: all features, selected features, and target-only. Our findings reveal highly variable effectiveness: while Chronos-2 achieves the best performance among foundation models (in zero-shot settings), the simple baseline frequently outperforms all foundation models in Singapore's stable climate, particularly for short-term horizons. Model architecture proves critical, with synergistic architectural implementations (TTM's channel-mixing, Chronos-2's grouped attention) consistently leveraging exogenous features, while other approaches show inconsistent benefits. Geographic context emerges as equally important, with foundation models demonstrating advantages primarily in variable climates. These results challenge assumptions about universal foundation model superiority and highlight the need for domain-specific models, specifically in the energy domain.
\end{abstract}

%

%
\section{Introduction}
Accurate electricity demand forecasting is critical for reliable power system operation, enabling better demand response programs, strategic infrastructure planning, and maintenance scheduling~\cite{ntiElectricityLoadForecasting2020}.
The increasing penetration of renewable technologies and smart grids~\cite{mirReviewElectricityDemand2020} necessitates greater focus on exogenous predictors---particularly weather variables---which become especially important as regions face more extreme weather events due to climate change~\cite{tanakaImpactWeatherChanges2022}.

Traditional forecasting methods range from: statistical approaches like ARIMA and exponential smoothing~\cite{deoliveiraForecastingMidlongTerm2018}; machine learning techniques like support vector machines and gradient-boosted trees~\cite{aguilarmadridShortTermElectricityLoad2021}; finally deep learning architectures like RNNs, LSTMs and transformers~\cite{tianDeepNeuralNetwork2018, nieTimeSeriesWorth2023}. 
More recently, time-series foundation models---large pre-trained neural architectures originally developed for language and vision---have emerged as a new paradigm. 
These models leverage information from diverse time-series contexts to enhance predictive capabilities, showing promise across various benchmarks. 
However, practical challenges remain in effectively incorporating exogenous features that have proven critical for forecast accuracy in traditional settings~\cite{chengPowerLSTMPowerDemand2017, zhuReviewProspectDatadriven2022}.

This paper empirically evaluates foundation models for electricity load forecasting with specific emphasis on exogenous feature integration.
We systematically compare the LSTM against several foundation models capable of modeling cross-channel relationships---MOIRAI, MOMENT, TinyTimeMixers (TTM), ChronosX and Chronos-2---across Singapore and Australian electricity markets at hourly and daily granularities. 
Our findings reveal that foundation models demonstrate mixed effectiveness in leveraging exogenous features, with performance varying significantly across models, forecasting horizons, and geographical contexts.

\section{Related Work}

In this section, we provide a brief overview of the field of electricity load forecasting, covering several paradigms of machine learning methods involved. 
In addition, we detail the time-series foundation models used in our analysis. 

\subsection{Local vs Global training paradigm}
\subsubsection{Local training.} Traditional approaches focus on statistical modeling and machine learning trained on localized datasets~\cite{hongProbabilisticElectricLoad2016, ntiElectricityLoadForecasting2020}. 
While simple and explainable, these methods struggle with exogenous feature integration, and thus lose out on precious contextual information when performing forecasts. 
Deep learning methods, particularly LSTMs, have addressed this limitation through non-linear modeling of past-future correlations and direct integration of exogenous features, establishing strong performance in electricity demand forecasting~\cite{chengPowerLSTMPowerDemand2017, tianDeepNeuralNetwork2018, bashirShortTermElectricity2022}.

\subsubsection{Global training.} 
Foundation models employ a global training paradigm using massive compilations of heterogeneous time-series datasets spanning energy, healthcare, and finance domains~\cite{goswamiMOMENTFamilyOpen2024}. 
Most include electricity demand data (e.g., ETT, Electricity) and common exogenous predictors (weather, economic data) in their training pools~\cite{goswamiMOMENTFamilyOpen2024, wooUnifiedTrainingUniversal2024}. 
Unlike local training, these models use lightweight fine-tuning on target datasets, reportedly achieving performance equal to or exceeding previous state-of-the-art methods despite minimal domain-specific training~\cite{meyerBenchmarkingTimeSeries2024a, asgharnezhadTimeSeriesFoundation2024}.

\subsection{Modeling cross-channel relationships}
Many foundation models adopt channel-independent architectures that segregate multivariate inputs into univariate channels for independent prediction~\cite{nieTimeSeriesWorth2023}. 
While this approach improves performance across diverse domains and multivariate outputs, it potentially overlooks exogenous variable relationships previously shown to be crucial for load forecasting~\cite{chengPowerLSTMPowerDemand2017, zhuReviewProspectDatadriven2022}.
Fortunately, several time-series foundation models address this limitation through various innovative methods:

\subsubsection{MOIRAI.} MOIRAI~\cite{wooUnifiedTrainingUniversal2024} incorporates multivariate interactions through a novel Any-variate Attention layer on flattened multivariate input sequences, assigning importance scores to variables. This enables cross-channel modeling in both zero-shot and fine-tuned settings.

\subsubsection{TinyTimeMixers (TTM).} TTM~\cite{ekambaramTinyTimeMixers2024} uses channel-independent pretraining but learns multivariate interactions during fine-tuning. The channel-mixer block in its TSMixer components~\cite{ekambaramTSMixerLightweightMLPMixer2023} is enabled during this phase, allowing explicit cross-channel correlation capture in the target domain.

\subsubsection{MOMENT.} MOMENT~\cite{goswamiMOMENTFamilyOpen2024} does not explicitly address multivariate modeling in its architecture, maintaining a channel-independent pretraining approach. However, it accepts multivariate inputs/outputs and may naively learn channel interdependencies through linear probing of the forecasting head on output embeddings.

\subsubsection{ChronosX.} ChronosX~\cite{arangoChronosXAdaptingPretrained2025} builds upon univariate foundation models, specifically Chronos~\cite{ansariChronosLearningLanguage2024}, by introducing an adapter module framework specifically catered for past and future covariates.

\subsubsection{Chronos-2.} Chronos-2~\cite{ansariChronos2UnivariateUniversal2025} transitions to an encoder-only architecture from the full encoder-decoder in its original version, and introduces Group Attention to innately offer multivariate and covariate-informed forecasting capabilities. 

Despite training on common multivariate electricity benchmarks, these models' effectiveness in leveraging inter-channel correlations has not been explicitly evaluated with commonly used exogenous data in high-frequency forecasting (e.g., weather and date-related variables).
We note that there have been several works benchmarking time series foundation models in electricity demand forecasting~\cite{asgharnezhadTimeSeriesFoundation2024, meyerBenchmarkingTimeSeries2024a}, but the models investigated were inherently univariate in nature. 
This gap, combined with prior work demonstrating the importance of such features in load forecasting, motivates our systematic investigation of the effectiveness of various multivariate modeling frameworks and methods in foundation models for electricity demand prediction.

\section{Methodology}
\begin{figure*}[ht]
    \centering
    \includegraphics[width=\linewidth]{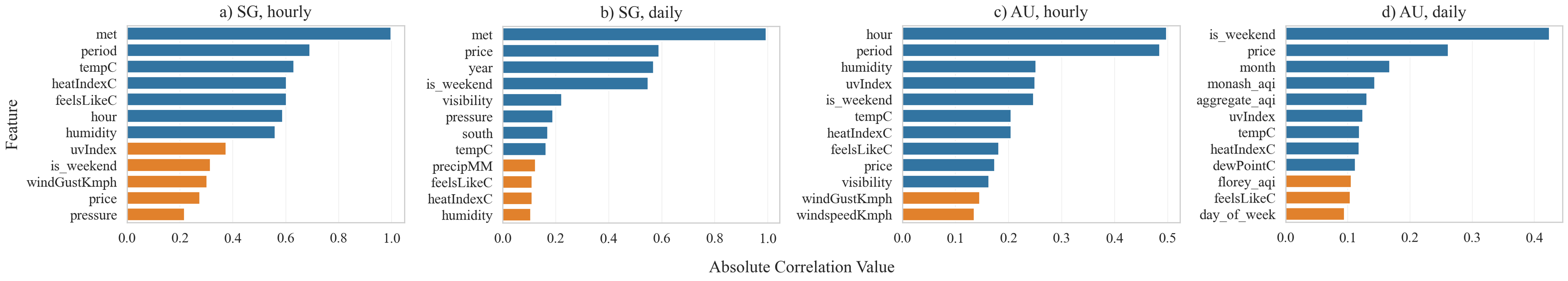}
    \caption{Top 12 absolute feature correlations by dataset. Selected features are colored in blue.}
    \label{fig:feature_correlations}
\end{figure*}

The following subsections outline the key components of our evaluation approach, from data collection and preprocessing to experimental configurations.

\subsection{Data Collection and Preprocessing}
We evaluate models on electricity demand data from Singapore (2016-2022) and Australia's ACT region (2015-2023), using three feature categories: demand, date-related, and weather features~\cite{chengPowerLSTMPowerDemand2017, christenExogenousDataLoad2020}.
Singapore's compact, climatically stable context contrasts with Australia's variable weather patterns and geographic spread~\cite{asgharnezhadTimeSeriesFoundation2024}. 

Demand variables were sourced from Singapore's Energy Market Authority and Australia's AEMO, aggregated to hourly and daily frequencies. 
Weather features came from World Weather Online, and air quality data from respective government portals. 
This yielded four datasets: Singapore and Australia at both hourly and daily granularities. 
After handling missing values via linear interpolation (continuous variables) and forward filling (categorical variables), we split data 60/20/20 for train/validation/test (Table \ref{tab:data_lengths}). The full set of variables can be found in Table \ref{tab:feature_sets} in the Appendix.

\subsubsection{Feature selection.} To assess exogenous variable impact, we tested three conditions: all features (~30 variables), selected features (7-10 variables), and target-only (no exogenous features). 
Selected features were identified via Spearman correlation with the target variable for simplicity, choosing features with notably higher absolute correlations while maintaining 7-10 features per dataset (Fig. \ref{fig:feature_correlations}).
Granger causality tests also confirmed the significance of these features' predictive capabilities at multiple lag intervals below the 512 context length.

\begin{table}[ht]
    \centering
    \begin{tabular}{lcccc}
    \toprule
    & \multicolumn{2}{c}{\textbf{Singapore}} & \multicolumn{2}{c}{\textbf{Australia}} \\
    & Hourly & Daily & Hourly & Daily \\
    \midrule
    Train & 36,821 & 1,535 & 47,333 & 1,972 \\
    Val/Test & 12,273 & 511 & 15,777 & 657 \\
    \bottomrule
    \end{tabular}
    \caption{Data lengths used in model training}
    \label{tab:data_lengths}
\end{table}

\subsection{Experimental configuration}
Aligning with past work~\cite{zhouInformerEfficientTransformer2021, wuAutoformerDecompositionTransformers2021, nieTimeSeriesWorth2023}, we processed the dataset into sliding window samples with a step size of 1.
Each window consists of the exogenous (if any) and target variables for 512 steps, as many foundation models had their context lengths fixed as such.
Forecast horizons were established as [1, 12, 24, 48, 72, 168, 336] for the hourly configuration (up to 14 days) and [1, 7, 14, 30, 60, 180, 365] for the daily configuration (up to 1 year).
Performance was evaluated using Mean Absolute Percentage Error (MAPE) across all prediction windows.

\subsection{Model selection and training}
We compared the following foundation models with explicit cross-channel correlation capabilities, each with differing approaches:
\begin{itemize}
    \item \textbf{MOIRAI:} Any-variate Attention layer enabling multivariate modeling in zero-shot and fine-tuned settings
    \item \textbf{TTM:} Channel-independent pretraining with explicit channel-mixing when fine-tuning
    \item \textbf{MOMENT:} Channel-independent pretraining with potential implicit learning through linear probing
    \item \textbf{ChronosX:} Lightweight training of adapter modules for time-series foundation models
    \item \textbf{Chronos2:} Group Attention across input channels to enable multivariate and covariate forecasting
\end{itemize}

Zero-shot experiments were conducted with MOIRAI (MOIRAI-ZS) and Chronos-2 (Chronos2-ZS) due to their architectures enabling inherent multivariate modeling. 
For the baseline, we used the LSTM with reversible instance normalization (RevIN,~\citealp{kimReversibleInstanceNormalization2021}) aligning with foundation models' inherent normalization techniques and reflecting RevIN's established superiority for time series forecasting~\cite{wooUnifiedTrainingUniversal2024, ekambaramTinyTimeMixers2024}.

\section{Results}
Our experiments reveal that foundation models demonstrate highly variable effectiveness in leveraging exogenous features for electricity demand forecasting. 
Performance varied significantly based across model architectures, forecasting horizons, and geographic contexts. 
Complete numerical results are presented in Table \ref{tab:hourly_results} (hourly) and Table \ref{tab:daily_results} (daily). 
Values in bold indicate overall best performance, while underlined values show best performance among foundation models only. 
$no\to all/sel$ represent percentage improvements from adding exogenous features (all or selected) relative to the condition without features.

\subsection{Hourly forecasting}
\begin{table*}[!t]
\renewcommand{\arraystretch}{1.2}
\centering
    \resizebox{\textwidth}{!}{
    \begin{tabular}{|c|c|c|c|c|c|c|c|c|c|c|c|c|c|c|c|}
    \hline
    \multirow{2}{*}{\textbf{Model}} & \multirow{2}{*}{\textbf{Config}} & \multicolumn{7}{c|}{\textbf{Singapore}} & \multicolumn{7}{c|}{\textbf{Australia}} \\
    \cline{3-16}
     &  & \textbf{1} & \textbf{12} & \textbf{24} & \textbf{48} & \textbf{72} & \textbf{168} & \textbf{336} & \textbf{1} & \textbf{12} & \textbf{24} & \textbf{48} & \textbf{72} & \textbf{168} & \textbf{336} \\
    \hline
    \multirow{5}{*}{MOMENT} & All features & 1.1559 & 1.5873 & 1.7436 & 1.9145 & 2.0042 & 2.2184 & 2.3746 & 3.4191 & 5.3057 & 5.6470 & 6.3157 & 6.6915 & 7.3090 & 8.2336 \\
     & Selected & 1.1394 & 1.5772 & 1.7471 & 1.9140 & 1.9995 & 2.2176 & 2.3748 & 3.5414 & 5.2727 & 5.6124 & 6.2943 & 6.6722 & 7.3062 & 8.2292 \\
     & No features & 1.2308 & 1.5311 & 1.7134 & 1.8869 & 1.9714 & 2.1962 & 2.3323 & 3.7788 & 5.2854 & 5.5962 & 6.1765 & 6.5355 & 7.0518 & 7.7727 \\
     & $no\to all$ (\%) & 6.08 & -3.67 & -1.76 & -1.46 & -1.66 & -1.01 & -1.81 & 9.52 & -0.38 & -0.91 & -2.25 & -2.39 & -3.65 & -5.93 \\
     & $no\to sel$ (\%) & 7.43 & -3.01 & -1.97 & -1.44 & -1.42 & -0.97 & -1.82 & 6.28 & 0.24 & -0.29 & -1.91 & -2.09 & -3.61 & -5.87 \\
    \hline
    \multirow{5}{*}{MOIRAI-ZS} & All features & 0.8477 & 1.5503 & 1.7809 & 2.0028 & 2.1561 & 2.5638 & 2.9777 & 2.2341 & 5.0071 & 5.2767 & 5.9743 & 6.4073 & 7.0242 & 7.9491 \\
     & Selected & 0.8251 & 1.5036 & 1.7191 & 1.9480 & 2.0562 & 2.3234 & 2.5802 & 2.2179 & 5.0240 & 5.2749 & 6.0438 & 6.5004 & 7.2535 & 8.0716 \\
     & No features & 0.8522 & 1.4803 & 1.6851 & 1.8901 & 1.9775 & 2.1203 & 2.2744 & 2.2010 & 4.8563 & 5.0228 & 5.5739 & 5.8517 & 6.3320 & 6.7897 \\
     & $no\to all$ (\%) & 0.52 & -4.73 & -5.69 & -5.96 & -9.03 & -20.92 & -30.92 & -1.50 & -3.10 & -5.06 & -7.18 & -9.49 & -10.93 & -17.08 \\
     & $no\to sel$ (\%) & 3.17 & -1.58 & -2.02 & -3.07 & -3.98 & -9.58 & -13.45 & -0.77 & -3.45 & -5.02 & -8.43 & -11.08 & -14.55 & -18.88 \\
    \hline
    \multirow{5}{*}{MOIRAI} & All features & 1.7673 & 1.9545 & 1.9945 & 2.1566 & 2.3229 & 2.5434 & 2.7003 & 5.1775 & 6.8604 & 6.7895 & 7.1641 & 7.3710 & 7.7246 & 8.1071 \\
     & Selected & 1.6954 & 1.9576 & 1.8074 & 1.9437 & 1.9914 & 2.1796 & 2.3707 & 5.5032 & 6.7540 & 6.5013 & 6.8689 & 7.0677 & 7.4590 & 7.8568 \\
     & No features & 2.3224 & 2.3527 & 2.2359 & 2.3391 & 2.3242 & 2.3908 & 2.5072 & 5.2999 & 7.0443 & 6.9671 & 7.4173 & 7.5902 & 7.9278 & 8.4139 \\
     & $no\to all$ (\%) & 23.90 & 16.93 & 10.79 & 7.80 & 0.06 & -6.38 & -7.70 & 2.31 & 2.61 & 3.71 & 3.41 & 2.89 & 2.56 & 3.65 \\
     & $no\to sel$ (\%) & 27.00 & 16.80 & 19.16 & 16.90 & 14.32 & 8.83 & 5.45 & -3.84 & 4.12 & 6.68 & 7.39 & 6.88 & 5.91 & 6.62 \\
    \hline
    \multirow{5}{*}{TTM} & All features & 0.6171 & 1.2226 & 1.5386 & 1.7258 & 1.8690 & 2.0623 & 2.3000 & 1.7874 & 3.8915 & 4.6505 & 5.2368 & 5.5151 & 6.1293 & 6.5062 \\
     & Selected & \underline{0.5680} & 1.3229 & 1.5946 & 1.8308 & 1.8942 & 2.1175 & 2.3009 & 1.8351 & 3.9453 & 4.4257 & 5.2555 & 5.6246 & 6.1656 & 6.5404 \\
     & No features & 0.7196 & 1.1126 & 1.3284 & 1.8396 & 1.7751 & 2.1443 & 2.1179 & 1.8453 & 4.6242 & 4.8058 & 5.3741 & 5.7109 & 6.2420 & 6.5393 \\
     & $no\to all$ (\%) & 14.24 & -9.89 & -15.82 & 6.18 & -5.29 & 3.82 & -8.60 & 3.13 & 15.85 & 3.23 & 2.55 & 3.43 & 1.81 & 0.51 \\
     & $no\to sel$ (\%) & 21.07 & -18.90 & -20.04 & 0.48 & -6.71 & 1.25 & -8.64 & 0.55 & 14.68 & 7.91 & 2.21 & 1.51 & 1.22 & -0.02 \\
    \hline
    \multirow{5}{*}{ChronosX\footnotemark[1]} & All features & 8.8133 & 1.7294 & 1.5241 & 1.7068 & 1.8753 & 2.3508 & 2.9258 & 13.1375 & 4.4663 & 4.9644 & 5.2503 & 5.3721 & 6.1273 & 6.6633 \\
     & Selected & 9.3958 & 1.5408 & 1.7832 & 1.9887 & 2.7329 & 3.1362 & \textbf{\underline{1.8707}} & 13.2211 & 4.4514 & 4.6518 & 5.2360 & 5.2825 & 6.7869 & \textbf{\underline{6.1259}} \\
     & No features & 0.6337 & 1.2513 & 1.4797 & 1.7623 & -- & -- & -- & 1.6951 & 4.1705 & 4.6712 & 5.3433 & -- & -- & -- \\
     & $no\to all$ (\%) & $<$-100 & -38.21 & -3.00 & 3.15 & -- & -- & -- & $<$-100 & -7.09 & -6.28 & 1.74 & -- & -- & -- \\
     & $no\to sel$ (\%) & $<$-100 & -23.14 & -20.51 & -12.85 & -- & -- & -- & $<$-100 & -6.74 & 0.42 & 2.01 & -- & -- & -- \\
    \hline
    \multirow{5}{*}{Chronos2-ZS} & All features & 0.6257 & 1.0717 & 1.3012 & 1.5215 & 1.6383 & 1.8861 & 2.0652 & 1.6846 & \textbf{\underline{3.5653}} & \textbf{\underline{4.0083}} & \textbf{\underline{4.6104}} & \textbf{\underline{4.9710}} & \textbf{\underline{5.6519}} & 6.1325 \\
     & Selected & 0.6493 & 1.0754 & 1.2987 & 1.5221 & 1.6384 & 1.8746 & 2.0454 & \textbf{\underline{1.6816}} & 3.5932 & 4.0327 & 4.6326 & 4.9935 & 5.6729 & 6.1630 \\
     & No features & 0.6033 & \underline{1.0690} & \underline{1.2944} & \underline{1.5169} & \textbf{\underline{1.6340}} & \textbf{\underline{1.8694}} & 2.0394 & 1.6856 & 3.6595 & 4.0676 & 4.6640 & 5.0035 & 5.6769 & 6.1611 \\
     & $no\to all$ (\%) & -3.71 & -0.25 & -0.53 & -0.31 & -0.26 & -0.89 & -1.26 & 0.06 & 2.58 & 1.46 & 1.15 & 0.65 & 0.44 & 0.46 \\
     & $no\to sel$ (\%) & -7.62 & -0.60 & -0.34 & -0.35 & -0.27 & -0.27 & -0.29 & 0.24 & 1.81 & 0.86 & 0.67 & 0.20 & 0.07 & -0.03 \\
    \hline
    \multirow{5}{*}{RevIN-LSTM} & All features & 0.5194 & 1.1541 & 1.3625 & 1.8485 & 1.9442 & 2.2409 & 2.3279 & 2.1697 & 6.2533 & 5.9250 & 6.9864 & 7.1922 & 8.0321 & 8.1399 \\
     & Selected & 0.5774 & 0.9498 & 1.1488 & 1.5032 & 1.7489 & 2.0768 & 2.2093 & 2.0894 & 6.2026 & 6.1513 & 7.2940 & 8.1764 & 7.5034 & 8.0626 \\
     & No features & \textbf{0.4401} & \textbf{0.8952} & \textbf{1.0374} & \textbf{1.4404} & 1.6479 & 2.0214 & 2.2074 & 2.1668 & 7.9578 & 6.7064 & 8.2311 & 9.0409 & 9.2151 & 9.1383 \\
     & $no\to all$ (\%) & -18.04 & -28.92 & -31.34 & -28.33 & -17.98 & -10.86 & -5.46 & -0.13 & 21.42 & 11.65 & 15.12 & 20.45 & 12.84 & 10.93 \\
     & $no\to sel$ (\%) & -31.21 & -6.09 & -10.74 & -4.36 & -6.13 & -2.74 & -0.09 & 3.57 & 22.06 & 8.28 & 11.39 & 9.56 & 18.57 & 11.77 \\
    \hline
    \end{tabular}
    }
\caption{Hourly MAPE (\%) Across Different Prediction Horizons}  
\label{tab:hourly_results}
\end{table*}

\subsubsection{Overall performance patterns.}
In hourly forecasting (Table \ref{tab:hourly_results}), Singapore consistently achieved lower MAPE values (0.44\%-3.14\%) compared to Australia's ACT region (1.68\%-9.22\%), reflecting the inherent predictability differences between a climatically stable city-state and a region with more variable weather patterns.
These stability differences are mirrored in how models responded to exogenous features: Australia demonstrated greater reliance on weather variables across most models and horizons.

Among foundation models, Chronos-2 consistently achieved the lowest MAPE values across both regions and most horizons, with TTM and ChronosX following closely behind.
Surprisingly, the baseline RevIN-LSTM without exogenous features frequently outperformed all foundation models in Singapore for short-term horizons (up to 48 hours), remaining highly competitive with all foundation models even at longer horizons.

\subsubsection{Exogenous feature impact.}
In Singapore, exogenous variables appear to be underutilized, with most models preferring univariate predictions over those with exogenous features.
Notable exceptions to this are: TTM, which displayed the most variable response with improvements up to 21.1\% but also deteriorations up to 20.0\% depending on horizon; and fine-tuned MOIRAI, showing sustained benefits across both contexts but remained the model with the poorest performance across the board.
Interestingly, MOMENT, MOIRAI and TTM all show improvements using exogenous features at the extreme short-term horizon of 1-hour ahead.
Conversely, ChronosX has drastically reduced performance when optimizing for 1-hour ahead forecasts, a pattern to be observed across all experiment configurations.

In Australia, exogenous variables were generally observed to be much more well received. 
With the exception of MOMENT and MOIRAI-ZS, all other models observed consistent improvements when employing exogenous information, with the largest increases observed by RevIN-LSTM.
Models like TTM and Chronos-2 slightly preferred the complete feature set, suggesting that higher-dimensional inputs are able to be handled by their architectures, especially when sufficient predictive signal is present. 
ChronosX observed a large discrepancy in its preference for feature sets, and in our limited experiments (both hourly and daily configurations) show a general preference towards the set of no features (i.e. univariate forecasting), despite its framework specifically targeting exogenous feature integration.

\subsection{Daily forecasting}
\begin{table*}[!t]
\renewcommand{\arraystretch}{1.2}
\centering
    \resizebox{\textwidth}{!}{
    \begin{tabular}{|c|c|c|c|c|c|c|c|c|c|c|c|c|c|c|c|}
    \hline
    \multirow{2}{*}{\textbf{Model}} & \multirow{2}{*}{\textbf{Config}} & \multicolumn{7}{c|}{\textbf{Singapore}} & \multicolumn{7}{c|}{\textbf{Australia}} \\
    \cline{3-16}
    & & \textbf{1} & \textbf{7} & \textbf{14} & \textbf{30} & \textbf{60} & \textbf{180} & \textbf{365} & \textbf{1} & \textbf{7} & \textbf{14} & \textbf{30} & \textbf{60} & \textbf{180} & \textbf{365} \\
    \hline
    \multirow{5}{*}{MOMENT} & All features & 2.3931 & 2.4467 & 2.4428 & 2.4530 & 2.4858 & 2.3961 & 2.8759 & 5.8517 & 6.4593 & 6.4876 & 6.6499 & 7.3059 & 7.2062 & 6.9865 \\
    & Selected & 2.0789 & 2.5056 & 2.6281 & 2.5579 & 2.6836 & 2.5689 & 2.6945 & 5.9011 & 6.3190 & 6.8066 & 7.2651 & 7.5403 & 7.3937 & 6.9715 \\
    & No features & 2.4604 & 2.2356 & 2.4133 & 2.4178 & 2.6764 & 2.4689 & 3.0068 & 6.2498 & 6.2679 & 6.7794 & 7.2817 & 7.5965 & 7.4636 & 7.0083 \\
    & $no\to all$ (\%) & 2.73 & -9.44 & -1.22 & -1.46 & 7.12 & 2.95 & 4.35 & 6.37 & -3.05 & 4.31 & 8.68 & 3.83 & 3.45 & 0.31 \\
    & $no\to sel$ (\%) & 15.51 & -12.08 & -8.90 & -5.80 & -0.27 & -4.05 & 10.39 & 5.58 & -0.82 & -0.40 & 0.23 & 0.74 & 0.94 & 0.53 \\
    \hline
    \multirow{5}{*}{MOIRAI-ZS} & All features & 2.9034 & 3.1261 & 3.1876 & 3.2896 & 3.4106 & 3.7686 & 4.1913 & 4.6037 & 5.6946 & 5.9411 & 6.8685 & 8.9175 & 10.5998 & 10.9706 \\
    & Selected & 2.8372 & 3.0825 & 3.1470 & 3.2254 & 3.3527 & 3.6969 & 4.3593 & 4.6034 & 5.6382 & 5.9319 & 6.8702 & 9.1592 & 11.5816 & 13.1086 \\
    & No features & 2.3738 & 2.2844 & 2.2065 & 2.2491 & 2.4125 & 9.3060 & 17.1858 & 3.6692 & 4.7645 & 4.8818 & 5.1334 & 5.8920 & 7.5973 & 10.5064 \\
    & $no\to all$ (\%) & -22.31 & -36.85 & -44.46 & -46.26 & -41.38 & 59.50 & 75.61 & -25.47 & -19.52 & -21.70 & -33.80 & -51.35 & -39.52 & -4.42 \\
    & $no\to sel$ (\%) & -19.52 & -34.94 & -42.62 & -43.41 & -38.97 & 60.27 & 74.63 & -25.46 & -18.34 & -21.51 & -33.83 & -55.45 & -52.44 & -24.77 \\
    \hline
    \multirow{5}{*}{MOIRAI} & All features & 3.1083 & 3.6777 & 3.7326 & 3.7882 & 3.7620 & 4.4256 & 4.6863 & 5.8583 & 6.3350 & 6.5615 & 6.8996 & 8.2427 & 7.0092 & 6.8606 \\
    & Selected & 3.4884 & 3.3769 & 3.3644 & 3.3498 & 3.3948 & 3.5651 & 4.1243 & 5.9031 & 7.5669 & 7.5477 & 8.1147 & 10.4745 & 11.0304 & 6.6605 \\
    & No features & 2.4872 & 2.3299 & 2.3437 & 2.6523 & 2.8972 & 4.0952 & 3.3135 & 6.0734 & 6.2201 & 6.4320 & 6.9962 & 7.8836 & 8.6827 & 7.7770 \\
    & $no\to all$ (\%) & -24.97 & -57.85 & -59.26 & -42.83 & -30.54 & -8.07 & -41.40 & 3.54 & -1.85 & -2.01 & 1.38 & -4.56 & 19.27 & 11.78 \\
    & $no\to sel$ (\%) & -40.25 & -44.94 & -43.55 & -26.30 & -17.18 & 12.95 & -24.47 & 2.80 & -21.65 & -17.35 & -15.99 & -32.86 & -27.04 & 14.36 \\
    \hline
    \multirow{5}{*}{TTM} & All features & 1.3888 & 1.8271 & 1.9588 & 2.1400 & 2.1839 & 2.1997 & 2.1101 & 3.3597 & 4.5054 & 4.7196 & 5.1151 & 5.2928 & 5.8941 & 6.1400 \\
    & Selected & 1.3823 & 1.9357 & 1.9641 & 2.1208 & 2.1538 & 2.1263 & 2.1759 & 3.4473 & 4.5077 & 4.7554 & 5.0708 & 5.3386 & 6.2522 & 6.1937 \\
    & No features & 1.4721 & 2.1536 & 2.1044 & 2.0509 & 2.1155 & 2.2359 & 2.4579 & 3.5117 & 4.7525 & 5.0861 & 5.3962 & 5.5892 & 6.0204 & 6.1715 \\
    & $no\to all$ (\%) & 5.65 & 15.16 & 6.92 & -4.35 & -3.24 & 1.62 & 14.15 & 4.33 & 5.20 & 7.21 & 5.21 & 5.30 & 2.10 & 0.51 \\
    & $no\to sel$ (\%) & 6.10 & 10.12 & 6.66 & -3.41 & -1.81 & 4.90 & 11.48 & 1.83 & 5.15 & 6.50 & 6.03 & 4.48 & -3.85 & -0.36 \\
    \hline
    \multirow{5}{*}{ChronosX} & All features & 3.6783 & 2.4309 & 2.1819 & 2.2052 & 2.1347 & 2.5229 & 2.8468 & 9.3096 & 4.8521 & 4.8489 & 6.1606 & 5.3482 & 8.0357 & 9.7992 \\
    & Selected & 3.6842 & 3.0246 & 2.2285 & 2.4553 & 2.8360 & 2.7466 & 2.3883 & 9.4224 & 5.1624 & 4.7180 & 5.8464 & 5.6033 & 7.9927 & 10.4607 \\
    & No features & 1.4405 & 1.8333 & 1.9616 & 2.0548 & 2.2067 & -- & -- & 3.3559 & 4.4365 & 4.6373 & 5.0461 & 5.8668 & -- & -- \\
    & $no\to all$ (\%) & $<$-100 & -32.59 & -11.23 & -7.32 & 3.26 & -- & -- & $<$-100 & -9.37 & -4.56 & -22.09 & 8.84 & -- & -- \\
    & $no\to sel$ (\%) & $<$-100 & -64.98 & -13.61 & -19.49 & -28.52 & -- & -- & $<$-100 & -16.36 & -1.74 & -15.86 & 4.49 & -- & -- \\
    \hline
    \multirow{5}{*}{Chronos2-ZS} & All features & \textbf{\underline{1.1637}} & \textbf{\underline{1.5631}} & \textbf{\underline{1.7121}} & \textbf{\underline{1.7620}} & \textbf{\underline{1.7882}} & \textbf{\underline{1.8166}} & 1.9290 & 3.0765 & 4.1035 & 4.3090 & 4.4961 & 4.6753 & \textbf{\underline{4.9923}} & 5.0874 \\
    & Selected & 1.1847 & 1.5813 & 1.7223 & 1.7659 & 1.7902 & 1.8193 & \textbf{\underline{1.9103}} & \textbf{\underline{3.0651}} & \textbf{\underline{4.0997}} & \textbf{\underline{4.2808}} & \textbf{\underline{4.4613}} & \textbf{\underline{4.6498}} & 5.0775 & 5.1905 \\
    & No features & 1.2298 & 1.6266 & 1.7470 & 1.8288 & 1.8871 & 2.2534 & 2.5959 & 3.1090 & 4.1149 & 4.3084 & 4.4790 & 4.7212 & 5.1790 & \textbf{\underline{5.0808}} \\
    & $no\to all$ (\%) & 5.37 & 3.90 & 2.00 & 3.65 & 5.24 & 19.38 & 25.69 & 1.05 & 0.28 & -0.01 & -0.38 & 0.97 & 3.61 & -0.13 \\
    & $no\to sel$ (\%) & 3.67 & 2.78 & 1.41 & 3.44 & 5.13 & 19.26 & 26.41 & 1.41 & 0.37 & 0.64 & 0.40 & 1.51 & 1.96 & -2.16 \\
    \hline
    \multirow{5}{*}{RevIN-LSTM} & All features & 1.5890 & 2.3602 & 2.4682 & 2.5202 & 2.4144 & 2.8486 & 3.2816 & 3.8296 & 5.1571 & 5.6156 & 5.7418 & 6.0120 & 6.3844 & 6.8098 \\
    & Selected & 1.5008 & 2.1176 & 2.4837 & 2.5993 & 2.1960 & 2.8300 & 5.1318 & 3.3222 & 5.2721 & 5.4698 & 5.7342 & 6.1166 & 7.1369 & 7.2684 \\
    & No features & 1.3524 & 1.9039 & 1.9286 & 3.4472 & 2.2262 & 2.7702 & 5.1294 & 3.4149 & 4.9316 & 5.7395 & 7.5295 & 7.5611 & 7.3372 & 7.3111 \\
    & $no\to all$ (\%) & -17.50 & -23.64 & -27.98 & -7.37 & -8.46 & -2.83 & 36.02 & -12.14 & -4.57 & 2.16 & 23.74 & 20.49 & 12.99 & 6.86 \\
    & $no\to sel$ (\%) & -10.98 & -10.93 & -28.78 & -10.74 & 1.35 & -2.16 & -0.05 & 2.71 & -6.90 & 4.70 & 23.84 & 19.10 & 2.73 & 0.58 \\
    \hline
    \end{tabular}
    }
    
\caption{Daily MAPE (\%) Across Different Prediction Horizons}
\label{tab:daily_results}
\end{table*}
\subsubsection{Overall performance patterns.}
Similar to the hourly setting, daily forecasting (Table \ref{tab:daily_results}) also observed Singapore consistently achieving lower MAPE values (1.35\%-5.13\%) compared to Australia's ACT region (3.07\%-13.11\%).
Chronos-2 maintained the strongest performance across the board in both contexts, followed by TTM.
ChronosX also showed competitive performances at horizons 60 and below, though it observed the same issue of drastic performance reductions with exogenous features at horizon 1.
In contrast, RevIN-LSTM has a much weaker showing in the daily configuration, only staying competitive in the shorter horizons and tapering off after horizon 30. 

\footnotetext[1]{ChronosX with no features, i.e. zero-shot Chronos, only supports forecasting up to a horizon of 64. Longer horizons, while possible, were not trained on and thus may give undesirable results.}

\subsubsection{Exogenous feature impact.}
Compared to the hourly setting, exogenous features appear to be much more utilized in the daily setting for both Singapore and Australia contexts, at least for specific models. 

In Singapore, exogenous variables were utilized to a great extent across many horizons by models like TTM (1.62\%-14.15\%) and Chronos-2 (1.41\%-26.41\%).
Other models also observed significant increases in performance at longer horizons, with MOIRAI-ZS observing more than 60\% increases, MOMENT with up to 10\% increases, and RevIN-LSTM observing a single 36.02\% increase with all features at horizon 365.
Fine-tuned MOIRAI no longer had performance increases with exogenous features unlike in the hourly setting, showing some variance in its ability to model cross-channel dependencies across different granularities.
Seemingly, Chronos-2 experiments utilizing the set of all features marginally outperformed those using the feature-selected set, a pattern also previously observed in Australia's hourly setting.

In Australia, like in the hourly setting, exogenous variables were generally more well-utilized. 
With the exception of MOIRAI, most models consistently observed increases in performance across various horizons, and of a notably higher proportion as compared to in Singapore. 
In contrast to the hourly setting, Chronos-2 performed marginally better when using the set of selected features.
Other models did not show consistent preferences for feature sets.

\section{Discussion}
The experimental results reveal complex patterns in how foundation models leverage exogenous features for electricity load forecasting. 
We organize our discussion by examining model-specific behaviors, patterns across contexts, and broader implications for the field.

\subsection{Model Architecture and Feature Utilization}
\subsubsection{Comparing Foundation Models.}
The varying effectiveness of exogenous features across foundation models can be largely attributed to their differing approaches to multivariate modeling.
With TTM and Chronos-2 observing the most positive responses to multivariate inputs, the architectural implementations of these models may thus pose the greatest benefits to the field of electricity demand forecasting.
TTM, with its paradigm of channel-mixing on fine-tuning, specifically tuned for its simple and unique MLP-like architecture, show the effectiveness of curated efforts in architectural design.
Chronos-2 presents an exceptionally compelling case for large-scale time-series modeling with exogenous variables, using an encoder-only transformer with simple twist: grouped attention across channels.
While attaining one of the best performances across all experimental settings, it also manages to further incorporate exogenous signals where other models may fail.
To take it a step further, this was done in a zero-shot setting, whereas other models required explicit, sometimes complex, fine-tuning processes. 

On the other hand, other models demonstrated increased variance in their outputs when processing exogenous features.
MOIRAI's any-variate attention layer, theoretically possess superior capabilities for capturing multivariate dependencies.
However, it observed deteriorations throughout most of its zero-shot settings when incorporating exogenous features.
This is likely due to the specificity of the any-variate attention in learning correlations only between variates that were exposed to the model during training.
MOMENT's inconsistency was not unexpected, as it was not originally designed to support modeling cross-channel relationships. 
Its architectural design relies entirely on the linear probing of output embeddings to learn cross-channel relationships, proving it an arduous task to learn correlations purely based off a downstream classification head alone. 
ChronosX proved to be the most unpredictable, with varying levels of improvement and deterioration with exogenous variables across contexts and horizons.
The simple attachment of adapter blocks to complex foundation models may appear to be an effective solution across models, but potentially experiences inhibited performance when it comes down to specific domains and contexts.

\subsubsection{Feature Utilization.}
The distinction between "all features" and "selected features" configurations also revealed intriguing insights into feature engineering requirements for foundation models.
The same models had varying preferences for feature sets in different settings, as observed in Chronos-2's preference for all features in Australia-hourly and Singapore-daily, while preferring selected features in Australia-daily instead.
When the data context provides sufficient supplementary information, as in Australia's more variable weather patterns, foundation models maybe be able to leverage a larger number of channels more effectively than conventional deep learning methods, owing to their large-scale pretraining.

\subsubsection{Encoder-only vs Decoder-only Transformers.}
Another notable pattern lies in the choice of underlying transformer architectures of current time-series foundation models. 
With the exception of TTM as an MLP variant, the models investigated in our study primarily rely on the encoder-only architecture.  
Whether or not it is a coincidence that only these models are able to perform forecasting with exogenous variables (by design) remains to be seen.
While encoder architectures are understandably adept at building a rich, latent representation of cross-channel dynamics to perform forecasts, time-series foundation modeling seems to be converging towards primarily decoder-only architectures for powerful auto-regressive forecasting, as observed in the release of models like TimesFM-2~\cite{dasDecoderonlyFoundationModel2024} and MOIRAI-2.0~\cite{wooUnifiedTrainingUniversal2024}.
Decoder-only models incorporating exogenous information may be a promising direction for future work, despite the potential complexity involved in high-dimensional multivariate forecasting in an autoregressive manner.

\subsubsection{Computational Efficiency.}
Although computational efficiency was not empirically investigated in our study, we briefly discuss general considerations to inform practical applicability. 
Foundation models typically incur substantially higher resource requirements than traditional deep learning approaches like the RevIN-LSTM baseline, both during fine-tuning and inference. 
Although TTM's lightweight MLP-based architecture offers a more favorable trade-off, transformer-based foundation models---particularly those employing multivariate attention mechanisms—often demand significantly greater memory and compute overhead. 
These costs scale with the number of input channels: attention-based approaches exhibit quadratic complexity with respect to the number of features, making high-dimensional covariate configurations increasingly expensive~\cite{wooUnifiedTrainingUniversal2024, ansariChronos2UnivariateUniversal2025}.

For real-time or near-real-time forecasting applications common in grid operations, such overhead may prove prohibitive without mitigation strategies. 
Potential approaches include more rigorous preliminary feature selection to reduce input dimensionality, or architectural innovations such as sparse attention mechanisms~\cite{royEfficientContentBasedSparse2021} and emerging alternatives such as state-space models~\cite{daoTransformersAreSSMs2024}.

\subsection{Geographic Context and Predictability}
Geographic context emerged as a critical determinant of both model performance and feature utility, revealing that the value of exogenous features is not universal but rather contingent on the underlying predictability of the electricity market.
Singapore's climatically stable environment resulted in consistently lower absolute errors, but also diminished benefits from exogenous features, with the simple RevIN-LSTM baseline outperforming foundation models in many cases. 
In contrast, Australia's higher climate variability made exogenous features more valuable across most models and horizons, with foundation models demonstrating stronger relative performance.

This pattern suggests that foundation models trained on heterogeneous datasets may be implicitly optimized for markets exhibiting high variability, where complex multivariate relationships provide genuine predictive value.
However, in more stable, predictable contexts where simple historical patterns and trends dominate, the additional modeling capacity of foundation models may introduce unnecessary complexity without proportionate benefits. 
The strong performance of RevIN-LSTM in Singapore indicates that for certain electricity markets, the sophisticated architectures and large-scale pretraining of foundation models may be overshooting the actual modeling requirements.
In these cases, training a smaller deep-learning model may thus prove to be more proportionately efficient than optimizing the entire set of weights in a large-scale foundation model.

\subsection{A Need for Foundation Models in Energy}

We surmise that the pretraining of foundation models on highly heterogeneous datasets spanning diverse domains may result in learned representations not optimally suited to any single domain. 
The electricity demand data in standard pretraining datasets may not adequately represent the full diversity of global electricity markets, particularly stable city-states like Singapore.
We acknowledge that the benefits of large-scale pretraining may be the most pronounced in low-data regimes where transfer learning provides critical inductive biases.
Our experiments expose models to substantial amounts of training data, which potentially reduces the relative advantage of pretraining.
In such situations, the value of foundation models becomes less clear, particularly when weighed against their computational requirements and increased complexity.

These findings point to a critical need for energy-domain foundation models that are pretrained specifically on diverse electricity and energy forecasting contexts, with explicit architectural attention to modeling well-established weather-demand relationships.
Encouragingly, recent work has begun moving toward domain specificity, with emerging foundation models for power systems and grids~\cite{tuPowerPMFoundationModel2024, hamannFoundationModelsElectric2024}, weather and climate forecasting~\cite{schmudePrithviWxCFoundation2024}, and benchmarks specifically tailored for energy forecasting across diverse contexts~\cite{wangBenchmarksCustomPackage2024}.
Building on this momentum, the field would benefit from systematic investigations into feature engineering, selection, and correlation modeling within energy-specific foundation models---determining not only whether these models can forecast accurately, but whether they can reliably harness the domain knowledge encoded in exogenous variables that the field has long recognized as critical for operational electricity demand forecasting.

\section{Conclusion}
This paper empirically evaluated time-series foundation models' effectiveness in leveraging exogenous features for electricity demand forecasting across Singaporean and Australian electricity markets at hourly and daily granularities. 
These findings challenge assumptions about the universal superiority of foundation models for domain-specific forecasting. 
Geographic context emerged as a critical moderating factor---foundation models showed stronger advantages in Australia's variable climate where weather features provided genuine predictive value, but introduced unnecessary complexity in Singapore's predictable environment where RevIN-LSTM dominated, particularly over short-term settings. 

This gap between the promise of large-scale pretraining and practical performance points to a need for energy-domain-specific foundation models with explicit architectural focus on cross-channel relationships.
Future research should investigate which mechanisms most reliably capture exogenous dependencies, 
how models can distinguish between stable and variable contexts, and 
what pretraining strategies optimize for effective covariate integration in operational electricity forecasting.

\bibliography{references}

@article{aguilarmadridShortTermElectricityLoad2021,
  title = {Short-{{Term Electricity Load Forecasting}} with {{Machine Learning}}},
  author = {Aguilar Madrid, Ernesto and Antonio, Nuno},
  year = 2021,
  month = feb,
  journal = {Information},
  volume = {12},
  number = {2},
  pages = {50},
  publisher = {Multidisciplinary Digital Publishing Institute},
  issn = {2078-2489},
  doi = {10.3390/info12020050},
  urldate = {2025-06-11},
  copyright = {http://creativecommons.org/licenses/by/3.0/},
  langid = {english}
}

@misc{ansariChronos2UnivariateUniversal2025,
  title = {Chronos-2: {{From Univariate}} to {{Universal Forecasting}}},
  shorttitle = {Chronos-2},
  author = {Ansari, Abdul Fatir and Shchur, Oleksandr and K{\"u}ken, Jaris and Auer, Andreas and Han, Boran and Mercado, Pedro and Rangapuram, Syama Sundar and Shen, Huibin and Stella, Lorenzo and Zhang, Xiyuan and Goswami, Mononito and Kapoor, Shubham and Maddix, Danielle C. and Guerron, Pablo and Hu, Tony and Yin, Junming and Erickson, Nick and Desai, Prateek Mutalik and Wang, Hao and Rangwala, Huzefa and Karypis, George and Wang, Yuyang and {Bohlke-Schneider}, Michael},
  year = 2025,
  month = oct,
  number = {arXiv:2510.15821},
  eprint = {2510.15821},
  primaryclass = {cs},
  publisher = {arXiv},
  doi = {10.48550/arXiv.2510.15821},
  urldate = {2025-10-22},
  archiveprefix = {arXiv}
}

@misc{ansariChronosLearningLanguage2024,
  title = {Chronos: {{Learning}} the {{Language}} of {{Time Series}}},
  shorttitle = {Chronos},
  author = {Ansari, Abdul Fatir and Stella, Lorenzo and Turkmen, Caner and Zhang, Xiyuan and Mercado, Pedro and Shen, Huibin and Shchur, Oleksandr and Rangapuram, Syama Sundar and Arango, Sebastian Pineda and Kapoor, Shubham and Zschiegner, Jasper and Maddix, Danielle C. and Wang, Hao and Mahoney, Michael W. and Torkkola, Kari and Wilson, Andrew Gordon and {Bohlke-Schneider}, Michael and Wang, Yuyang},
  year = 2024,
  month = nov,
  number = {arXiv:2403.07815},
  eprint = {2403.07815},
  primaryclass = {cs},
  publisher = {arXiv},
  doi = {10.48550/arXiv.2403.07815},
  urldate = {2025-10-08},
  archiveprefix = {arXiv}
}

@misc{arangoChronosXAdaptingPretrained2025,
  title = {{{ChronosX}}: {{Adapting Pretrained Time Series Models}} with {{Exogenous Variables}}},
  shorttitle = {{{ChronosX}}},
  author = {Arango, Sebastian Pineda and Mercado, Pedro and Kapoor, Shubham and Ansari, Abdul Fatir and Stella, Lorenzo and Shen, Huibin and Senetaire, Hugo and Turkmen, Caner and Shchur, Oleksandr and Maddix, Danielle C. and {Bohlke-Schneider}, Michael and Wang, Yuyang and Rangapuram, Syama Sundar},
  year = 2025,
  month = mar,
  number = {arXiv:2503.12107},
  eprint = {2503.12107},
  primaryclass = {cs},
  publisher = {arXiv},
  doi = {10.48550/arXiv.2503.12107},
  urldate = {2025-06-11},
  archiveprefix = {arXiv}
}

@misc{asgharnezhadTimeSeriesFoundation2024,
  type = {{{SSRN Scholarly Paper}}},
  title = {Time {{Series Foundation Models}} for {{Load Forecasting}}: {{A Comprehensive Analysis}}},
  shorttitle = {Time {{Series Foundation Models}} for {{Load Forecasting}}},
  author = {Asgharnezhad, Hamzeh and Tabarisaadi, Pegah and Khosravi, Abbas and Quan, Hao},
  year = 2024,
  month = dec,
  number = {5067419},
  eprint = {5067419},
  publisher = {Social Science Research Network},
  address = {Rochester, NY},
  doi = {10.2139/ssrn.5067419},
  urldate = {2025-04-28},
  archiveprefix = {Social Science Research Network},
  langid = {english}
}

@article{bashirShortTermElectricity2022,
  title = {Short Term Electricity Load Forecasting Using Hybrid Prophet-{{LSTM}} Model Optimized by {{BPNN}}},
  author = {Bashir, Tasarruf and Haoyong, Chen and Tahir, Muhammad Faizan and Liqiang, Zhu},
  year = 2022,
  month = nov,
  journal = {Energy Reports},
  volume = {8},
  pages = {1678--1686},
  issn = {2352-4847},
  doi = {10.1016/j.egyr.2021.12.067},
  urldate = {2025-06-09}
}

@inproceedings{chengPowerLSTMPowerDemand2017,
  title = {{{PowerLSTM}}: {{Power Demand Forecasting Using Long Short-Term Memory Neural Network}}},
  shorttitle = {{{PowerLSTM}}},
  booktitle = {Advanced {{Data Mining}} and {{Applications}}},
  author = {Cheng, Yao and Xu, Chang and Mashima, Daisuke and Thing, Vrizlynn L. L. and Wu, Yongdong},
  editor = {Cong, Gao and Peng, Wen-Chih and Zhang, Wei Emma and Li, Chengliang and Sun, Aixin},
  year = 2017,
  pages = {727--740},
  publisher = {Springer International Publishing},
  address = {Cham},
  doi = {10.1007/978-3-319-69179-4_51},
  isbn = {978-3-319-69179-4},
  langid = {english}
}

@inproceedings{christenExogenousDataLoad2020,
  title = {Exogenous {{Data}} for {{Load Forecasting}}: {{A Review}}:},
  shorttitle = {Exogenous {{Data}} for {{Load Forecasting}}},
  booktitle = {Proceedings of the 12th {{International Joint Conference}} on {{Computational Intelligence}}},
  author = {Christen, Ram{\'o}n and Mazzola, Luca and Denzler, Alexander and Portmann, Edy},
  year = 2020,
  pages = {489--500},
  publisher = {{SCITEPRESS - Science and Technology Publications}},
  address = {Budapest, Hungary},
  doi = {10.5220/0010213204890500},
  urldate = {2025-06-11},
  isbn = {978-989-758-475-6},
  langid = {english}
}

@misc{daoTransformersAreSSMs2024,
  title = {Transformers Are {{SSMs}}: {{Generalized Models}} and {{Efficient Algorithms Through Structured State Space Duality}}},
  shorttitle = {Transformers Are {{SSMs}}},
  author = {Dao, Tri and Gu, Albert},
  year = 2024,
  month = may,
  number = {arXiv:2405.21060},
  eprint = {2405.21060},
  primaryclass = {cs},
  publisher = {arXiv},
  doi = {10.48550/arXiv.2405.21060},
  urldate = {2025-12-08},
  archiveprefix = {arXiv}
}

@misc{dasDecoderonlyFoundationModel2024,
  title = {A Decoder-Only Foundation Model for Time-Series Forecasting},
  author = {Das, Abhimanyu and Kong, Weihao and Sen, Rajat and Zhou, Yichen},
  year = 2024,
  month = apr,
  number = {arXiv:2310.10688},
  eprint = {2310.10688},
  primaryclass = {cs},
  publisher = {arXiv},
  doi = {10.48550/arXiv.2310.10688},
  urldate = {2025-06-02},
  archiveprefix = {arXiv}
}

@article{deoliveiraForecastingMidlongTerm2018,
  title = {Forecasting Mid-Long Term Electric Energy Consumption through Bagging {{ARIMA}} and Exponential Smoothing Methods},
  author = {{de Oliveira}, Erick Meira and Cyrino Oliveira, Fernando Luiz},
  year = 2018,
  month = feb,
  journal = {Energy},
  volume = {144},
  pages = {776--788},
  issn = {0360-5442},
  doi = {10.1016/j.energy.2017.12.049},
  urldate = {2025-06-11}
}

@misc{ekambaramTinyTimeMixers2024,
  title = {Tiny {{Time Mixers}} ({{TTMs}}): {{Fast Pre-trained Models}} for {{Enhanced Zero}}/{{Few-Shot Forecasting}} of {{Multivariate Time Series}}},
  shorttitle = {Tiny {{Time Mixers}} ({{TTMs}})},
  author = {Ekambaram, Vijay and Jati, Arindam and Dayama, Pankaj and Mukherjee, Sumanta and Nguyen, Nam H. and Gifford, Wesley M. and Reddy, Chandra and Kalagnanam, Jayant},
  year = 2024,
  month = nov,
  number = {arXiv:2401.03955},
  eprint = {2401.03955},
  primaryclass = {cs},
  publisher = {arXiv},
  doi = {10.48550/arXiv.2401.03955},
  urldate = {2025-10-08},
  archiveprefix = {arXiv}
}

@inproceedings{ekambaramTSMixerLightweightMLPMixer2023,
  title = {{{TSMixer}}: {{Lightweight MLP-Mixer Model}} for {{Multivariate Time Series Forecasting}}},
  shorttitle = {{{TSMixer}}},
  booktitle = {Proceedings of the 29th {{ACM SIGKDD Conference}} on {{Knowledge Discovery}} and {{Data Mining}}},
  author = {Ekambaram, Vijay and Jati, Arindam and Nguyen, Nam and Sinthong, Phanwadee and Kalagnanam, Jayant},
  year = 2023,
  month = aug,
  eprint = {2306.09364},
  primaryclass = {cs},
  pages = {459--469},
  doi = {10.1145/3580305.3599533},
  urldate = {2025-06-02},
  archiveprefix = {arXiv}
}

@misc{garzaTimeGPT12024,
  title = {{{TimeGPT-1}}},
  author = {Garza, Azul and Challu, Cristian and {Mergenthaler-Canseco}, Max},
  year = 2024,
  month = may,
  number = {arXiv:2310.03589},
  eprint = {2310.03589},
  primaryclass = {cs},
  publisher = {arXiv},
  doi = {10.48550/arXiv.2310.03589},
  urldate = {2025-06-02},
  archiveprefix = {arXiv}
}

@misc{goswamiMOMENTFamilyOpen2024,
  title = {{{MOMENT}}: {{A Family}} of {{Open Time-series Foundation Models}}},
  shorttitle = {{{MOMENT}}},
  author = {Goswami, Mononito and Szafer, Konrad and Choudhry, Arjun and Cai, Yifu and Li, Shuo and Dubrawski, Artur},
  year = 2024,
  month = oct,
  number = {arXiv:2402.03885},
  eprint = {2402.03885},
  primaryclass = {cs},
  publisher = {arXiv},
  doi = {10.48550/arXiv.2402.03885},
  urldate = {2025-10-08},
  archiveprefix = {arXiv}
}

@misc{hamannFoundationModelsElectric2024,
  title = {Foundation {{Models}} for the {{Electric Power Grid}}},
  author = {Hamann, Hendrik F. and Brunschwiler, Thomas and Gjorgiev, Blazhe and Martins, Leonardo S. A. and Puech, Alban and Varbella, Anna and Weiss, Jonas and {Bernabe-Moreno}, Juan and Mass{\'e}, Alexandre Blondin and Choi, Seong and Foster, Ian and Hodge, Bri-Mathias and Jain, Rishabh and Kim, Kibaek and Mai, Vincent and Mirall{\`e}s, Fran{\c c}ois and Montigny, Martin De and {Ramos-Lea{\~n}os}, Octavio and Supr{\^e}me, Hussein and Xie, Le and Youssef, El-Nasser S. and Zinflou, Arnaud and Belyi, Alexander J. and Bessa, Ricardo J. and Bhattarai, Bishnu Prasad and Schmude, Johannes and Sobolevsky, Stanislav},
  year = 2024,
  month = nov,
  number = {arXiv:2407.09434},
  eprint = {2407.09434},
  primaryclass = {cs},
  publisher = {arXiv},
  doi = {10.48550/arXiv.2407.09434},
  urldate = {2025-11-01},
  archiveprefix = {arXiv}
}

@article{hongProbabilisticElectricLoad2016,
  title = {Probabilistic Electric Load Forecasting: {{A}} Tutorial Review},
  shorttitle = {Probabilistic Electric Load Forecasting},
  author = {Hong, Tao and Fan, Shu},
  year = 2016,
  month = jul,
  journal = {International Journal of Forecasting},
  volume = {32},
  number = {3},
  pages = {914--938},
  issn = {0169-2070},
  doi = {10.1016/j.ijforecast.2015.11.011},
  urldate = {2025-06-09}
}

@inproceedings{kimReversibleInstanceNormalization2021,
  title = {Reversible {{Instance Normalization}} for {{Accurate Time-Series Forecasting}} against {{Distribution Shift}}},
  booktitle = {International {{Conference}} on {{Learning Representations}}},
  author = {Kim, Taesung and Kim, Jinhee and Tae, Yunwon and Park, Cheonbok and Choi, Jang-Ho and Choo, Jaegul},
  year = 2021,
  month = oct,
  urldate = {2025-06-05},
  langid = {english}
}

@misc{meyerBenchmarkingTimeSeries2024a,
  title = {Benchmarking {{Time Series Foundation Models}} for {{Short-Term Household Electricity Load Forecasting}}},
  author = {Meyer, Marcel and Zapata, David and Kaltenpoth, Sascha and M{\"u}ller, Oliver},
  year = 2024,
  month = oct,
  number = {arXiv:2410.09487},
  eprint = {2410.09487},
  primaryclass = {cs},
  publisher = {arXiv},
  doi = {10.48550/arXiv.2410.09487},
  urldate = {2025-04-28},
  archiveprefix = {arXiv}
}

@article{mirReviewElectricityDemand2020,
  title = {A {{Review}} of {{Electricity Demand Forecasting}} in {{Low}} and {{Middle Income Countries}}: {{The Demand Determinants}} and {{Horizons}}},
  shorttitle = {A {{Review}} of {{Electricity Demand Forecasting}} in {{Low}} and {{Middle Income Countries}}},
  author = {Mir, Aneeque A. and Alghassab, Mohammed and Ullah, Kafait and Khan, Zafar A. and Lu, Yuehong and Imran, Muhammad},
  year = 2020,
  month = jan,
  journal = {Sustainability},
  volume = {12},
  number = {15},
  pages = {5931},
  publisher = {Multidisciplinary Digital Publishing Institute},
  issn = {2071-1050},
  doi = {10.3390/su12155931},
  urldate = {2024-10-15},
  copyright = {http://creativecommons.org/licenses/by/3.0/},
  langid = {english}
}

@misc{nieTimeSeriesWorth2023,
  title = {A {{Time Series}} Is {{Worth}} 64 {{Words}}: {{Long-term Forecasting}} with {{Transformers}}},
  shorttitle = {A {{Time Series}} Is {{Worth}} 64 {{Words}}},
  author = {Nie, Yuqi and Nguyen, Nam H. and Sinthong, Phanwadee and Kalagnanam, Jayant},
  year = 2023,
  month = mar,
  series = {{{ICLR}} '23},
  number = {arXiv:2211.14730},
  eprint = {2211.14730},
  publisher = {arXiv},
  doi = {10.48550/arXiv.2211.14730},
  urldate = {2024-10-21},
  archiveprefix = {arXiv}
}

@article{ntiElectricityLoadForecasting2020,
  title = {Electricity Load Forecasting: A Systematic Review},
  shorttitle = {Electricity Load Forecasting},
  author = {Nti, Isaac Kofi and Teimeh, Moses and {Nyarko-Boateng}, Owusu and Adekoya, Adebayo Felix},
  year = 2020,
  month = sep,
  journal = {Journal of Electrical Systems and Information Technology},
  volume = {7},
  number = {1},
  pages = {13},
  issn = {2314-7172},
  doi = {10.1186/s43067-020-00021-8},
  urldate = {2025-06-09}
}

@article{royEfficientContentBasedSparse2021,
  title = {Efficient {{Content-Based Sparse Attention}} with {{Routing Transformers}}},
  author = {Roy, Aurko and Saffar, Mohammad and Vaswani, Ashish and Grangier, David},
  year = 2021,
  month = feb,
  journal = {Transactions of the Association for Computational Linguistics},
  volume = {9},
  pages = {53--68},
  issn = {2307-387X},
  doi = {10.1162/tacl_a_00353},
  urldate = {2025-12-08}
}

@misc{schmudePrithviWxCFoundation2024,
  title = {Prithvi {{WxC}}: {{Foundation Model}} for {{Weather}} and {{Climate}}},
  shorttitle = {Prithvi {{WxC}}},
  author = {Schmude, Johannes and Roy, Sujit and Trojak, Will and Jakubik, Johannes and Civitarese, Daniel Salles and Singh, Shraddha and Kuehnert, Julian and Ankur, Kumar and Gupta, Aman and Phillips, Christopher E. and Kienzler, Romeo and Szwarcman, Daniela and Gaur, Vishal and Shinde, Rajat and Lal, Rohit and Silva, Arlindo Da and Diaz, Jorge Luis Guevara and Jones, Anne and Pfreundschuh, Simon and Lin, Amy and Sheshadri, Aditi and Nair, Udaysankar and Anantharaj, Valentine and Hamann, Hendrik and Watson, Campbell and Maskey, Manil and Lee, Tsengdar J. and Moreno, Juan Bernabe and Ramachandran, Rahul},
  year = 2024,
  month = sep,
  number = {arXiv:2409.13598},
  eprint = {2409.13598},
  primaryclass = {cs},
  publisher = {arXiv},
  doi = {10.48550/arXiv.2409.13598},
  urldate = {2025-11-01},
  archiveprefix = {arXiv}
}

@article{tanakaImpactWeatherChanges2022,
  title = {The Impact of Weather Changes on the Supply and Demand of Electric Power and Wholesale Prices of Electricity in {{Germany}}},
  author = {Tanaka, Kenta and Matsumoto, Ken'ichi and Keeley, Alexander Ryota and Managi, Shunsuke},
  year = 2022,
  month = sep,
  journal = {Sustainability Science},
  volume = {17},
  number = {5},
  pages = {1813--1825},
  publisher = {Springer Japan},
  issn = {1862-4057},
  doi = {10.1007/s11625-022-01219-7},
  urldate = {2025-06-09},
  copyright = {2022 The Author(s), under exclusive licence to Springer Japan KK, part of Springer Nature},
  langid = {english}
}

@article{tianDeepNeuralNetwork2018,
  title = {A {{Deep Neural Network Model}} for {{Short-Term Load Forecast Based}} on {{Long Short-Term Memory Network}} and {{Convolutional Neural Network}}},
  author = {Tian, Chen and Ma, Jian and Zhang, Chunhong and Zhan, Peida},
  year = 2018,
  journal = {Energies},
  volume = {11},
  number = {12},
  doi = {10.3390/en11123493},
  urldate = {2024-10-21},
  langid = {english}
}

@misc{tuPowerPMFoundationModel2024,
  title = {{{PowerPM}}: {{Foundation Model}} for {{Power Systems}}},
  shorttitle = {{{PowerPM}}},
  author = {Tu, Shihao and Zhang, Yupeng and Zhang, Jing and Fu, Zhendong and Zhang, Yin and Yang, Yang},
  year = 2024,
  month = oct,
  number = {arXiv:2408.04057},
  eprint = {2408.04057},
  primaryclass = {cs},
  publisher = {arXiv},
  doi = {10.48550/arXiv.2408.04057},
  urldate = {2025-06-25},
  archiveprefix = {arXiv}
}

@misc{wangBenchmarksCustomPackage2024,
  title = {Benchmarks and {{Custom Package}} for {{Energy Forecasting}}},
  author = {Wang, Zhixian and Wen, Qingsong and Zhang, Chaoli and Sun, Liang and Krannichfeldt, Leandro Von and Pan, Shirui and Wang, Yi},
  year = 2024,
  month = oct,
  number = {arXiv:2307.07191},
  eprint = {2307.07191},
  primaryclass = {cs},
  publisher = {arXiv},
  doi = {10.48550/arXiv.2307.07191},
  urldate = {2025-10-17},
  archiveprefix = {arXiv}
}

@misc{wooUnifiedTrainingUniversal2024,
  title = {Unified {{Training}} of {{Universal Time Series Forecasting Transformers}}},
  author = {Woo, Gerald and Liu, Chenghao and Kumar, Akshat and Xiong, Caiming and Savarese, Silvio and Sahoo, Doyen},
  year = 2024,
  month = may,
  series = {{{ICML}} '24},
  number = {arXiv:2402.02592},
  eprint = {2402.02592},
  primaryclass = {cs},
  publisher = {arXiv},
  doi = {10.48550/arXiv.2402.02592},
  urldate = {2025-07-02},
  archiveprefix = {arXiv}
}

@inproceedings{wuAutoformerDecompositionTransformers2021,
  title = {Autoformer: {{Decomposition Transformers}} with {{Auto-Correlation}} for {{Long-Term Series Forecasting}}},
  shorttitle = {Autoformer},
  booktitle = {Advances in {{Neural Information Processing Systems}}},
  author = {Wu, Haixu and Xu, Jiehui and Wang, Jianmin and Long, Mingsheng},
  year = 2021,
  volume = {34},
  pages = {22419--22430},
  publisher = {Curran Associates, Inc.},
  urldate = {2024-10-21}
}

@article{zhouInformerEfficientTransformer2021,
  title = {Informer: {{Beyond Efficient Transformer}} for {{Long Sequence Time-Series Forecasting}}},
  shorttitle = {Informer},
  author = {Zhou, Haoyi and Zhang, Shanghang and Peng, Jieqi and Zhang, Shuai and Li, Jianxin and Xiong, Hui and Zhang, Wancai},
  year = 2021,
  month = may,
  journal = {Proceedings of the AAAI Conference on Artificial Intelligence},
  volume = {35},
  number = {12},
  pages = {11106--11115},
  issn = {2374-3468},
  doi = {10.1609/aaai.v35i12.17325},
  urldate = {2024-10-21},
  copyright = {Copyright (c) 2021 Association for the Advancement of Artificial Intelligence},
  langid = {english}
}

@article{zhuReviewProspectDatadriven2022,
  title = {Review and Prospect of Data-Driven Techniques for Load Forecasting in Integrated Energy Systems},
  author = {Zhu, Jizhong and Dong, Hanjiang and Zheng, Weiye and Li, Shenglin and Huang, Yanting and Xi, Lei},
  year = 2022,
  month = sep,
  journal = {Applied Energy},
  volume = {321},
  pages = {119269},
  issn = {0306-2619},
  doi = {10.1016/j.apenergy.2022.119269},
  urldate = {2025-10-31}
}

\appendix
\section{A \quad Detailed Experiment Settings}
\subsubsection{Model configurations.}
All models were trained or fine-tuned with standardized configurations whenever possible to minimize the impact of hyperparameter selection and ensure fair comparison. 
Unless otherwise specified, foundation models were run and fine-tuned with the default parameters from their respective GitHub repositories.
Mean-Squared Error (MSE) was used as the training loss function throughout. 
The Adam optimizer was used, with the learning rate set to 0.001.
For TTM, MOMENT and RevIN-LSTM, the maximum number of epochs for training was set to 100 with an early stopping patience of 10.
Experiments were run on a single A40 GPU, with the batch size set to 16 to accommodate larger models. 
The full sets of features used in our experiments can be found below in Table \ref{tab:feature_sets}.

The following versions of the selected foundation models were used: 
\begin{itemize}
    \item \texttt{AutonLab/MOMENT-1-large}
    \item \texttt{Salesforce/moirai-1.0-R-large}
    \item \texttt{ibm-granite/granite-timeseries-ttm-r2}
    \item \texttt{amazon/chronos-t5-small} (as ChronosX base)
    \item \texttt{amazon/chronos-2}
\end{itemize}
For \texttt{TTM-R2}, we used the most optimal revisions for each forecasting horizon, as their context and prediction lengths are fixed: \texttt{main} for horizons 96 and below, \texttt{512-192-r2} for horizons between 97 and 192, \texttt{512-336-r2} for horizons between 193 and 336, and \texttt{512-720-r2} for horizons beyond 336. 
Chronos-2 was not fine-tuned due to a combination of time constraints and fine-tuning being an experimental feature.

\subsubsection{Data sources.} 

Target demand and related features were sourced from Singapore's Energy Market Authority\footnote{https://www.ema.gov.sg/resources/statistics/half-hourly-system-demand-data} 
and Australia's AEMO\footnote{https://aemo.com.au/energy-systems/electricity/national-electricity-market-nem/data-nem/aggregated-data}, aggregated to hourly and daily frequencies. 
Weather features came from World Weather Online\footnote{https://www.worldweatheronline.com/}, 
and air quality data from the respective government portals\footnote{https://data.gov.sg/datasets}\footnote{https://www.data.act.gov.au/Environment/Air-Quality-Monitoring-Data/94a5-zqnn/about\_data}. 

\subsubsection{Other foundation models.} While MOIRAI has released its weights for its updated \texttt{moirai-2.0-R-small} variant, the architecture has changed significantly and it is unclear how inter-channel correlations are being modeled, so we choose to omit it in our experiments.
TimeGPT~\cite{garzaTimeGPT12024}, while popular as a commercial model and its ability to model cross-channel relationships, the closed-source nature of the model makes it unsuitable for investigation.

\begin{table*}[!h]
    \centering
    \begin{tabularx}{\textwidth}{llp{1.8cm}p{1.5cm}p{2.5cm}Xp{2cm}}
        \toprule
        \textbf{Context} & \textbf{Frequency} & \textbf{Config.} & \textbf{Demand Features} & \textbf{Date Features} & \textbf{Weather Features} & \textbf{Air Quality Features} \\
        \midrule

        \multirow{24}{*}{Singapore}
        & \multirow{12}{*}{Hourly}
        & All features
        & met\footnotemark, price 
        & year, month, day, hour, period, day\_of\_week, is\_weekend 
        & humidity, tempC, heatIndexC, precipMM, windspeedKmph, winddirDegree, windGustKmph, weatherDesc, visibility, pressure, cloudcover, dewPointC, uvIndex, feelsLikeC 
        & north, south, east, west, central, aggregatePSI \\ \cmidrule{3-7}
        
        & & Feature-selected
        & met
        & period, hour 
        & tempC, heatIndexC, feelsLikeC, humidity 
        & - \\ \cmidrule{3-7}

        & & Historical only
        & - 
        & - 
        & - 
        & - \\ \cmidrule{2-7}
        
        & \multirow{12}{*}{Daily}
        & All features
        & met, price 
        & year, month, day, day\_of\_week, is\_weekend 
        & humidity, tempC, heatIndexC, precipMM, windspeedKmph, winddirDegree, windGustKmph, weatherDesc, visibility, pressure, cloudcover, dewPointC, uvIndex, feelsLikeC 
        & north, south, east, west, central, aggregatePSI \\ \cmidrule{3-7}
        
        & & Feature-selected 
        & price 
        & year, is\_weekend 
        & visibility, pressure, tempC 
        & south \\ \cmidrule{3-7}
        
        & & Historical only 
        & - 
        & - 
        & - 
        & - \\
        
        \midrule
        
        \multirow{26}{*}{Australia}
        & \multirow{13}{*}{Hourly}
        & All features
        & price 
        & year, month, day, hour, period, day\_of\_week, is\_weekend 
        & humidity, tempC, heatIndexC, precipMM, windspeedKmph, winddirDegree, windGustKmph, weatherDesc, visibility, pressure, cloudcover, dewPointC, uvIndex, feelsLikeC 
        & civic\_aqi, florey\_aqi, monash\_aqi, aggregate\_aqi \\ \cmidrule{3-7}
        
        & & Feature-selected 
        & price 
        & hour, period, is\_weekend 
        & humidity, uvIndex, tempC, heatIndexC, feelsLikeC, visibility 
        & - \\ \cmidrule{3-7}
        
        & & Historical only 
        & - 
        & - 
        & - 
        & - \\ \cmidrule{2-7}
        
        & \multirow{13}{*}{Daily}
        & All features
        & price 
        & year, month, day, day\_of\_week, is\_weekend 
        & humidity, tempC, heatIndexC, precipMM, windspeedKmph, winddirDegree, windGustKmph, weatherDesc, visibility, pressure, cloudcover, dewPointC, uvIndex, feelsLikeC 
        & civic\_aqi, florey\_aqi, monash\_aqi, aggregate\_aqi \\ \cmidrule{3-7}
        
        & & Feature-selected 
        & price 
        & is\_weekend, month 
        & uvIndex, tempC, heatIndexC, dewPointC 
        & monash\_aqi, aggregate\_aqi \\ \cmidrule{3-7}
        
        & & Historical only 
        & - 
        & - 
        & - 
        & - \\
        
        \bottomrule
    \end{tabularx}
\caption{Features by Context, Frequency, and Configuration}
\label{tab:feature_sets}
\end{table*}
\footnotetext{Actual Demand met by all \textit{Generation Registered Facilities}. Differs from target \textit{System Demand}.}

\end{document}